\documentclass[11pt]{article}

\usepackage{ACL2023}

\usepackage{times}
\usepackage{latexsym}

\usepackage{multirow}

\usepackage{colortbl}

\usepackage{arydshln}
\usepackage{graphicx}
\usepackage{booktabs}
\usepackage{subcaption}
\usepackage{multirow}
\usepackage{rotating}
\usepackage{amsmath}
\usepackage[normalem]{ulem}
\usepackage{lscape}
\usepackage{ctable}
\usepackage{amsfonts}
\usepackage{quoting}
\usepackage[belowskip=-10pt,skip=0.333\baselineskip]{caption}
\usepackage[T1]{fontenc}
\usepackage{array}
\newcolumntype{C}[1]{>{\centering\let\newline\\\arraybackslash\hspace{0pt}}m{#1}}
\newcolumntype{P}[1]{>{\centering\arraybackslash}p{#1}}

\usepackage{makecell}

\setcellgapes{3pt}
\makegapedcells

\usepackage{hhline}

\usepackage[utf8]{inputenc}

\usepackage{microtype}
\usepackage{hyperref}

\usepackage{inconsolata}
\title{Math Word Problem Solving by Generating Linguistic Variants of\\Problem Statements} 

\author{\bf Syed Rifat Raiyan, 
{\bf Md. Nafis Faiyaz,}
{\bf Shah Md. Jawad Kabir,}\\ 
{\bf Mohsinul Kabir,}
{\bf Hasan Mahmud,}
{\bf Md Kamrul Hasan}\\
Systems and Software Lab (SSL)\\ Department of Computer Science and Engineering\\
Islamic University of Technology, Dhaka, Bangladesh\\
\texttt{\{rifatraiyan, nafisfaiyaz, jawadkabir, hasan, hasank\}@iut-dhaka.edu}\\
}
\begin{document}
\maketitle
\begin{abstract}
The art of mathematical reasoning stands as a fundamental pillar of intellectual progress and is a central catalyst in cultivating human ingenuity. Researchers have recently published a plethora of works centered around the task of solving Math Word Problems (MWP) --- a crucial stride towards general AI. These existing models are susceptible to dependency on shallow heuristics and spurious correlations to derive the solution expressions. In order to ameliorate this issue, in this paper, we propose a framework for MWP solvers based on the generation of linguistic variants of the problem text. The approach involves solving each of the variant problems and electing the predicted expression with the majority of the votes. We use DeBERTa (Decoding-enhanced BERT with disentangled attention) as the encoder to leverage its rich textual representations and enhanced mask decoder to construct the solution expressions. Furthermore, we introduce a challenging dataset, \textsc{ParaMAWPS}, consisting of paraphrased, adversarial, and inverse variants of selectively sampled MWPs from the benchmark \textsc{Mawps} dataset. We extensively experiment on this dataset along with other benchmark datasets using some baseline MWP solver models. We show that training on linguistic variants of problem statements and voting on candidate predictions improve the mathematical reasoning and robustness of the model. We make our code and data publicly available.
\end{abstract}

\section{Introduction}
\label{sec:intro}
Math word problem solving is a long-standing research problem in Artificial General Intelligence (AGI) and a lot of studies about this topic, from both industry and academia, have been published recently. A typical Math Word Problem (MWP) takes the form of a written narrative that articulates a problem scenario and poses a question regarding one or more unknown quantities. A language model capable of solving such problems has to translate the human-readable problem statement to a valid mathematical expression that can be evaluated to obtain the numeric answer. An example of a classic MWP is portrayed in Table \ref{tab:tab1}, where the reader is asked to infer the revenue of a boutique shop.
    \begin{table}
    \centering
    \footnotesize
    \begin{tabular}{l}
    \hline
    \hline
    \begin{tabular}[c]{@{}l@{}}\textbf{Problem:} 69 handbags are sold for \$13 each. There are a\\total of 420 handbags in a boutique and the remaining ha-\\ndbags are sold for \$7 each. How much did the boutique\\earn after selling all the handbags?\end{tabular} \\ \hline
    \textbf{Expression:} $x = 69\times13+(420-69)\times7$                        \\ \hline
    \textbf{Solution:} 3354                                                        \\ \hline
    \hline
    \\
    \end{tabular}
    \caption{An example of a Math Word Problem.}
    \label{tab:tab1}
    \end{table}
    Such problems are generally found in math textbooks of $1^{st}$ to $8^{th}$ grade students and are easily solvable by humans with decent mathematical aptitude. 
    
    A lot of challenges manifest while designing an automated system for solving these problems \cite{zhang2019gap,sundaram2022nlp}. The primary challenge is to understand the quantities in the problem and capture their complex mathematical interconnections from a linear textual sequence written in natural language. There exists a diverse range of MWPs with differing difficulty levels, \textit{i.e.}, varying numbers of unknown values, and depth of the relationships between quantities, which require good mathematical reasoning ability to solve. Furthermore, the absence of crucial information and the presence of irrelevant information in the problem statements proves to be quite a challenge for the solver models \cite{patel2021nlp}. Other challenges include learning to tackle the chronological and temporal ambiguities of the events happening in the problem statements and dealing with MWPs that significantly differ from the training set in terms of semantic and syntactic structure.
    
    To address the problem outlined in Table \ref{tab:tab1}, a competent MWP solver model would need to possess the ability to associate the quantity, \textit{i.e.}, $69$ handbags, with its price attribute of $\$13$, and understand the relative arithmetic order by deriving $351$ remaining handbags, \textit{i.e.}, $420 - 69$, before associating the price attribute of $\$7$.
    A lot of psychological studies have been done on how human beings learn to solve mathematical problems and improve their aptitude \cite{piaget2013child,jbpetersoniq,sheripsychology}. The frontier of research involving MWP solving is considered a momentous step towards the apogee of AGI \cite{bubeck2023sparks} and so researchers have dedicated their efforts to replicating these complex cognitive patterns exhibited by human beings within the frameworks of AI models. The existing methods that are considered strong baselines for MWP solving can be demonstrably shown to use shallow heuristics to solve many of the MWPs in the benchmark datasets \cite{patel2021nlp} creating a faux impression of their mathematical reasoning capability. To account for this limitation, in this paper --- 
    \begin{itemize}
    \setlength\itemsep{0em}
        \item We propose a framework for solving simple math word problems by generating paraphrased linguistic variants of the input problem statement using OpenAI's latest Generative Pre-trained Transformer (GPT-3) \cite{brown2020language} models, namely \textit{text-davinci-003} and \textit{gpt-3.5-turbo}. The problem statement variants along with the original problem text then undergo the appropriate pre-processing steps and are fed to an MWP solver model with a DeBERTa-based encoder and Enhanced Mask decoder. 

        \item We also generate a large, augmented version of the \textsc{Mawps} \cite{koncel2016mawps} dataset, namely \textsc{ParaMAWPS} (\textbf{Para}phrased \textbf{MA}th \textbf{W}ord \textbf{P}roblem \textbf{S}olving Repository), as a challenging dataset by the introduction of paraphrased structural variations of almost all categories of problems, but emphasizing more on the categories that the strong baseline models find difficult to solve.
    \end{itemize}
    DeBERTa (Decoding-enhanced BERT with disentangled attention) \cite{he2020deberta} is currently one of the most popular pretrained language models due to its effectiveness in achieving state-of-the-art results on a variety of natural language processing tasks, including language translation, text classification, and question answering.
    In our work, we find that the DeBERTa model achieves value accuracies of $63.5\%$ and $91.0\%$ on the \textsc{Svamp} dataset \cite{patel2021nlp} and the \textsc{Mawps} dataset \cite{koncel2016mawps} respectively. It falls behind the current SOTA accuracy of \textsc{RoBERTa-DeductReasoner} \cite{jie2022learning} by a slight margin of $1 \pm 0.20\%$ on the \textsc{Mawps} dataset, but exceeds its accuracy of $47.3 \pm 0.20\%$ on the \textsc{Svamp} dataset. Our code and data are publicly available at \url{https://github.com/Starscream-11813/Variational-Mathematical-Reasoning}

\section{Problem Formulation}

A Math Word Problem $S$ is a sequence of word tokens and numeric values, where the $V_S = \{v_1,\dots,v_m\}$ denotes the word tokens in $S$ and the set $n_S = \{n_1,\dots,n_l\}$ denotes the set of numeric quantities in $S$. The set of word tokens $V_S$ consists of entities such as names of people, objects, units, and rates while the set of quantities $n_S$ consists of the numerical amount relevant to those entities. 

The goal of an MWP solver model is to map $S$ to a valid mathematical expression $E$, consisting of the quantities in $(n_S \cup C)$, where $C$ is a set of constants, and the fundamental mathematical operators $O = \{+,-,\times,\div\}$, which can be evaluated to obtain the correct answer.
\section{Literature Review}
\subsection{Math Word Problem Solving}
\subsubsection{Preliminary Works}
The dawn of research on MWP solving was in the mid-1960s \cite{feigenbaum1963computers,bobrow1964natural}. 
\textit{Rule-based methods} \cite{fletcher1985understanding,bakman2007robust,yuhui2010frame} are chronologically some of the earliest approaches to solving MWPs. They use a set of manually hard-coded rules about the language they are analyzing to find out regularities in the data. \textit{Statistical methods} \cite{kushman2014learning,hosseini2014learning,roy2015reasoning,zhou2015learn,mitra2016learning,liang2016tag,liang-etal-2016-meaning} use generic ML classifiers to extract the entities, quantities, and operators from the problem statement and infer the numeric answer with simple logic. \textit{Tree-based methods} \cite{koncel2015parsing,roy2016solving,roy2016equation,roy2017unit} utilize the inherent binary tree-like structure of expressions/equations.
Other primitive categories of approaches that have now been rendered somewhat obsolete are \textit{Parsing-based methods} \cite{shi2015automatically,zou2019text2math}, \textit{Similarity-based methods} \cite{huang2016well}, and \textit{Template-based methods} \cite{kushman2014learning,zhou2015learn,roy2016equation,upadhyay2016learning,huang2017learning}. 
\subsubsection{Deep Learning-based Methods}
Currently, the landscape of Deep learning models for the MWP solving task is primarily comprised of five distinct paradigms, \textsc{Seq2Seq}-based, \textsc{Seq2Tree}-based, \textsc{Graph2Tree}-based, \textit{complex relation extraction-based}, and \textit{Large Language Model (LLM) prompt-based} approaches, each of which has demonstrated remarkable levels of performance and efficacy.

\citet{wang2017deep} were the pioneers of introducing deep learning to solve MWPs with their proposed \textsc{Seq2Seq} model. To improve the \textsc{Seq2Seq} model, researchers resorted to alternative strategies, such as reinforcement learning techniques \cite{wang2018mathdqn,huang2018neural}, using dense problem representation \cite{mishra2018equgener}, adopting template-based methodologies \cite{wang2019template}, and incorporating group attention mechanisms \cite{li2019modeling}. \citet{xie2019goal} were the progenitors of the novel Goal-driven Tree-Structured (\textsc{Gts}) model, designed to generate expression trees using the tree-based decoder in order to imitate the goal-driven problem-solving approach of humans. The use of this tree decoder along with pre-trained language models, such as BERT \cite{jdevlin2018bert}, BART \cite{lewis2019bart}, RoBERTa \cite{liu2019roberta}, as the encoder in some of the \textsc{Seq2Tree} approaches \cite{liu2019tree,shen2020solving,wu2020knowledge,lin2021hms,shen2021generate,liang2021mwp,liangmwp,li2021seeking,xiong2022self} brought about substantial performance improvements over the previous \textsc{Seq2Seq} methods. \citet{cao2021bottom} devised a directed acyclic graph (\textsc{Seq2DAG}) model of the equations for the purpose of extracting the expression. \citet{zhang2020teacher} incorporated the idea of Knowledge Distillation (KD) \cite{hinton2015distilling} in their proposed model where the \textit{teacher network} is pre-trained to guide the learning behaviors of the \textit{student networks}. 
\citet{yu2021improving} introduced 2 types of encoders in their model.
\citet{hong2021learning} modified the work of \citet{xie2019goal} by incorporating a symbolic reasoning based \textit{Learning-by-fixing} (\textsc{Lbf}) framework. 
\citet{qin2021neural} proposed a model that performs 4 auxiliary tasks, Number Prediction, Commonsense Constant Prediction, Program Consistency Checking, and Duality Exploitation, to integrate different levels of symbolic constraints. 
\citet{huang2021recall} attempted to emulate human-like analogical learning in their proposed memory-augmented model. \textsc{Graph2Tree}-based approaches \cite{zhang2020graph,li2020graph} fused the merits of Graph-based Transformer \cite{yun2019graph,cai2020graph} encoders with multiple Graph Convolutional Network (multiGCN) modules \cite{kipf2016semi}, and tree-based decoders to solve MWPs. \citet{chatterjee2021weakly} introduced a weakly supervised approach for MWP solving. \citet{li2021seeking} introduced a contrastive learning approach with pattern divergence to solve MWPs. \citet{jie2022learning} formulated the MWP solving task as a complex relation extraction problem and leveraged explainable deductive reasoning techniques to iteratively construct the target equations. 

With the advent of LLMs, many innovative prompt-based methods \cite{shao2022chaining,li2022advance,wang2022self,pi2022reasoning,chen2022program,liang2023let} of solving MWPs that capitalize on the models' exceptional few-shot learning capability came into the limelight and demonstrated good performance across numerous benchmark datasets. \citet{cobbe2021training} used verifiers with their GPT-3 \cite{brown2020language} model.
Although LLMs excel at natural language understanding and have serendipitous emergent reasoning abilities \cite{yang2023harnessing}, they are still lackluster in complex reasoning tasks \cite{huang2022towards}. Numerous studies on complex reasoning tasks have empirically demonstrated that the approach of fine-tuning smaller models is superior \cite{ho2022large} to adopting LLM prompting techniques like Chain of Thought (CoT) prompting \cite{wei2022chain}.

\subsection{Paraphrasing}
Paraphrase generation has garnered significant attention from various NLP approaches, encompassing rule-based methods \cite{mckeown1980paraphrasing,meteer1988strategies}, data-driven techniques \cite{madnani2010generating}, linguistic translation methods \cite{bannard2005paraphrasing,barzilay2001extracting,prakash2016neural} that leverage bilingual corpora for iterative refinement by alternating back and forth between the languages \cite{madnani2010generating,prakash2016neural,mallinson-etal-2017-paraphrasing}. \citet{witteveen2019paraphrasing} demonstrated the superiority of LLMs like GPT-3 over the preceding methods in the paraphrasing task.

Accordingly, our work attempts to leverage the strengths of GPT-3 to generate a more linguistically diverse pool of problem statements to fine-tune a relatively smaller DeBERTa solver model on the downstream task of MWP solving which falls under the rubric of complex reasoning tasks.

\section{Methodology}
Figure-\ref{fig:proposedmodel} in Appendix-\ref{sec:appendix} shows an overview of our proposed architecture. Given a problem statement $S$, we prompt the paraphraser model to generate $k$ linguistic variants of $S$ which are, $S_1, S_2, \dots, S_k$. These $k$ variant problems along with the seed problem $S$ consists of quantities that are tagged appropriately using quantity tags. Each of the $k+1$ text sequences is then tokenized and the content embeddings $H$ and positional embeddings $P$ of the tokens are fed to the DeBERTa model. The disentangled self-attention mechanism of DeBERTa's encoder utilizes $H$ and $P$ to generate the output $H_{output}$, which is a contextual representation of the content of each problem statement. $H_{output}$, along with the relative positional embeddings $P$ and absolute positional embeddings $I$ of each of the problem statements are used by the Transformer layers of Enhanced Mask Decoder (EMD) of DeBERTa to generate the $k+1$ predicted equations $E_1, E_2, \dots, E_{k+1}$. These equations are then simplified and the equation that is predicted the most number of times is elected as the final prediction of the model. This majority voting module is used only during the validation/testing phase and for inference. During the training phase, the $k+1$ problem statements are deemed as stand-alone training samples and the Negative Log-Likelihood loss (NLLLoss) is calculated using the predicted equations and the ground-truth equation. Consequently, if the training set of the dataset used to train the model consists of $n$ samples, it is as if the model is trained with $(k+1)\times n = kn + n$ samples. The knowledge points gathered after being trained on an extra $kn$ samples contributes to the robustness of the model.
\subsection{Paraphrasing Model}
The task of correctly reformulating a Math Word Problem statement requires a good level of language understanding which is not present in its entirety in rule-based and data-driven methods of paraphrasing rendering them unsuitable in this case. These methods frequently yield incorrect, incoherent, and grammatically inaccurate linguistic variations; sometimes even leaving out crucial numerical information. Accordingly, we choose \textit{text-davinci-003} and \textit{gpt-3.5-turbo}, two GPT-3 models from OpenAI, as the paraphrasing models. GPT-3 (Generative Pre-trained Transformer 3) \cite{brown2020language} is a large language model with 175 billion parameters, that is capable of performing a wide range of natural language processing tasks, including paraphrasing a given sentence. Upon being prompted, it restates a given problem statement in different words while still maintaining the original meaning. To select the most appropriate paraphrase, GPT-3 uses a scoring mechanism that evaluates the semantic similarity between the original sentence and each of the generated paraphrases. The model assigns a higher score to paraphrases that are more similar in meaning to the input sentence, based on its understanding of the context and the relationships between the words. It also allows users to customize the level of complexity and the style of writing in the paraphrased version. We generate \(k\) variants of the original problem text by prompting the model. 
\subsubsection{Prompts and System Task Description}
The prompts that we use for accomplishing our linguistic variant generation task are,
        \begin{itemize}
        \small
            \item \textbf{\texttt{system role} Task Description} ---\\
            \texttt{You are a Math Word Problem rephraser that generates variations of math word problem statements.}
            \item \textbf{\texttt{user role} Prompts} ---
            \begin{itemize}
                \item \texttt{Generate }$k_1$\texttt{ paraphrased variations of the problem by changing the sentence structure.}
                \item \texttt{Generate }$k_2$\texttt{ paraphrased variations of the problem by changing the named entities and objects.}
                \item \texttt{Generate }$k_3$\texttt{ paraphrased variations of the problem with irrelevant numerical information.}
            \end{itemize}
        \end{itemize}
        Here, the total number of linguistic variants of a problem, $k = k_1 + k_2 + k_3$ and $5 \leq k \leq 15$.

A detailed discussion on the types of problem variations is delineated in Section-\ref{sec:experiment}.

\subsection{Quantity Tagging}
All the quantities (written either numerically or in words) in every single variant of the problem along with the original problem itself, are tagged with unique quantity tags using RegEx and a Python script which is provided in our GitHub repository (see Section-\ref{sec:intro}). This quantity tagging step ensures that the same quantity is present in both the input as well as in the output. The quantity-tagged tokens have their own content and positional embeddings. For example, if the problem statement is,
\begin{quoting}
\textit{``Melanie picked 4 plums, Dan picked 9 plums, and Sally picked 3 plums from the plum tree. How many plums were picked in total?"}
\end{quoting}
then the quantity-tagged version of the problem statement is,
\begin{quoting}
\textit{``Melanie picked \textsc{[Q1]} plums, Dan picked \textsc{[Q2]} plums, and Sally picked \textsc{[Q3]} plums from the plum tree. How many plums were picked in total?"}
\end{quoting}
We use this quantity tagging for the ground truth equation's quantities as well. 

\subsection{Encoder}
We use the pre-trained language model DeBERTa (\textbf{D}ecoding \textbf{e}nhanced \textbf{BERT} with disentangled \textbf{a}ttention). DeBERTa is a newly developed neural language model by \citet{he2020deberta} that is based on the Transformer architecture. It boasts a significant advancement over previous state-of-the-art (SOTA) pre-trained language models (PLMs) due to the incorporation of two novel techniques. The first technique is a disentangled attention mechanism and the second technique is an enhanced mask decoder. Together, these techniques make DeBERTa a highly effective PLM that outperforms its predecessors on a wide range of NLP downstream tasks.
\subsubsection{Disentangled Attention}
Contrary to BERT, which utilizes a vector representation for each word in the input layer by summing its content and position embeddings, in DeBERTa, every word is represented by two separate vectors that encode its content and position individually. The attention scores between words are computed using separate matrices that are disentangled based on the content and relative position of each word. This design choice is based on the observation that the attention weight between a pair of tokens is influenced by both their content and in tandem their relative positions. This especially holds paramount importance for the task of MWP solving as the relative positions of certain keywords in the problem statements dictate the solution.

To represent a token $x_i$ located at a specific position $i$ within a given sequence, it employs two distinct vectors, $H_i$ and $P_{i|j}$, which are respectively the content and relative positional representation vectors of $x_i$ with respect to a token $x_j$ at position $j$. The inter-token attention weights between $x_i$ and $x_j$ can be broken down into four constituent components,
\begin{equation}
\label{eq:eq1}
    \begin{split}
    A_{ij} &= \langle H_i, P_{i|j} \rangle \times \langle H_j, P_{j|i} \rangle^\top \\
           &= \underbrace{H_iH_j^\top}_{C2C} + \underbrace{H_iP_{j|i}^\top}_{C2P} + \underbrace{P_{i|j}H_j^\top}_{P2C} + \underbrace{P_{i|j}P_{j|i}^\top}_{\substack{P2P\\(omitted)}}
    \end{split}
\end{equation}
where, the four disentangled matrix attention scores represent their contents and positions as \textit{content-to-content (C2C)}, \textit{content-to-position (C2P)}, \textit{position-to-content (P2C)}, and \textit{position-to-position (P2P)}. The P2P portion of (\ref{eq:eq1}) is somewhat rendered obsolete since DeBERTa uses relative positional embedding which is why no useful information can be extracted from it.

The self-attention mechanism described by \citet{vaswani2017attention} has 3 parameters, $Q$ (Query), $K$ (Key), and $V$ (Value). The non-contextual embedding that is being contextualized at any point requests for information from its surrounding tokens within the context window and that is represented by the query token, and the tokens that the model pays attention to are the key tokens. 
\begin{equation}
\label{eq:eq2}
    \begin{split}
    Q_c &= HW_{c_Q}, K_c = HW_{c_K}, V_c = HW_{c_V} \\
    Q_r &= PW_{r_Q}, K_r = PW_{r_K}
    \end{split}
\end{equation}
where, $W_{c_Q} \in \mathbb{R}^{d \times d}$, $W_{c_K} \in \mathbb{R}^{d \times d}$, $W_{c_V} \in \mathbb{R}^{d \times d}$ are the projection weight matrices for the projected content vectors $Q_c$, $K_c$, $V_c$ respectively. Similarly, $W_{r_Q} \in \mathbb{R}^{d \times d}$ and $W_{r_K} \in \mathbb{R}^{d \times d}$ play the role of projection matrices for the projected relative position vectors $Q_r$ and $K_r$. The metric to calculate the relative distance between tokens $x_i$ and $x_j$ is,
\begin{equation}
    \label{eq:eq3}
    \delta(i,j) =
    \begin{cases}
        0, & \text{if $i - j \leq -k$}\\
        2k-1, & \text{if $i-j \geq k$}\\
        i-j+k, & \text{otherwise}
    \end{cases}
\end{equation}
which implies, $\delta(i,j) \in [0,2k)$. Each element $\bar{A}_{ij}$ of the attention matrix $\bar{A}$ denotes the attention score from token $x_i$ to the token $x_j$ and is computed using the vectors defined in (\ref{eq:eq2}) in the following manner,
\begin{equation}
    \label{eq:eq4}
    \bar{A}_{ij} = \underbrace{Q_i^cK_j^{c\top}}_{C2C} + \underbrace{Q_i^cK_{\delta(i,j)}^{r\top}}_{C2P} + \underbrace{K_j^cQ_{\delta(j,i)}^{r\top}}_{P2C}
\end{equation}
The attention score is yielded using the dot-product of the query and key in the formula to let the model have an idea of how similar the key is to the query. The output of the self-attention mechanism, which is denoted by $H_{output} \in \mathbb{R}^{N \times d}$ is,
\begin{equation}
    \label{eq:eq5}
    H_{output} = \mathbf{softmax}\left(\frac{\bar{A}}{\sqrt{3d}}\right)V_c
\end{equation}
The result of the dot-product is normalized by dividing with $\sqrt{3d}$ to avoid very hard softmax with small gradients, which is especially required for training stability in the case of large-scale PLMs \cite{vaswani2017attention,he2020deberta}.

\subsection{Decoder}
\citet{he2020deberta} postulates that the premature integration of absolute positions, which is employed by BERT \cite{jdevlin2018bert} in its decoding phase, could potentially impede the model's ability to acquire adequate knowledge of relative positions. With this as the justification, DeBERTa, being a model that was pre-trained using MLM (Masked Language Modeling), uses the absolute positions of the tokens in the penultimate layer, right before the softmax layer during the masked token prediction in its decoding phase. This enables all the Transformer layers in the decoder to work with the relative positional information without the susceptibility of hampering the learning process of the model. Since the absolute positions of the tokens in a sentence highly influence the nuanced understanding of the sentence's semantic and syntactic structure, and extracting information from only the relative positions isn't sufficient, the absolute positions are incorporated in the tail-end of the pipeline in the case of DeBERTa. This is why DeBERTa's decoding module is dubbed an Enhanced Mask Decoder (EMD) and it demonstrably outperforms the decoder counterparts of its predecessor PLMs \cite{he2020deberta}.

\subsection{Majority Voting}
Since there can be multiple valid equations for a single MWP, each of the $k+1$ predictions from the decoder, $E_1, E_2 \dots, E_{k+1}$, is simplified to a reduced normal form using the python package \texttt{sympy}\footnote{\url{https://www.sympy.org/en/index.html}}. These $k+1$ simplified predictions, $E'_1, E'_2 \dots, E'_{k+1}$, are then counted and the prediction that is the most frequent or that is yielded the most number of times is elected as the final answer of the whole solver model. It is to be noted that this voting mechanism is used only during the testing/validation phases or during inference.
    \begin{equation}
    \setlength{\belowdisplayskip}{0pt}
    \label{eq:eq6}
        E^* \leftarrow \operatorname*{argmax}_{E'} \mathbf{Votes}(E'_{i}); \quad i=1, 2, \dots, k+1
    \end{equation}

\section{Experiment}
\label{sec:experiment}
\subsection{Data Acquisition}
We introduce a new large-scale dataset, namely \textsc{ParaMAWPS} (\textbf{Para}phrased \textbf{MA}th \textbf{W}ord \textbf{P}roblem \textbf{S}olving Repository), consisting of 16,278 single equation MWPs. It is generated as a by-product of using one of the most commonly-used English MWP datasets, \textsc{Mawps} \cite{koncel2016mawps} which consists of a total of 2,373 problems, and the paraphraser model. We save the generated paraphrased variants of selectively sampled problems of \textsc{Mawps} and also manually include inverse versions of the problems to create our dataset. The dataset contains all the problems from the original \textsc{Mawps} dataset as well as paraphrased versions of some of the more challenging problems within \textsc{Mawps}, hence the name, \textsc{ParaMawps}. The samples are manually checked for correctness by 3 undergraduate students. By generating variations of some of the more difficult problems, we intend to increase familiarity of challenging concepts found within those problems to any model trained over this data, as well as more thoroughly challenge existing models trained on datasets that do not provide said complexity at an equal or higher density.
We generate $k$ problems from each seed problem in the dataset, adding up to a total of $k+1$ problems, where $5\leq k \leq 16$. Each of the $k$ generated problems will be a variation on the original that will feature several changes to the problem text. We generate 4 types of variations of each seed problem (see Table-\ref{tab:variations} in Appendix-\ref{sec:appendix}).
\begin{itemize}
    \item \textbf{Changed phrase order} ---
        Variations with the order of the phrases being changed facilitate a break from the standard problem statement template where quantities are generally given before the question formulation. Having a changed ordering of phrases makes apriori question formulations more common.
    \item \textbf{Changed object and entity names} ---
        Object and entity names are altered with interchangeable alternatives (names, synonyms) in problem variations to prevent fixation on elements of the problem mostly agnostic to the process of solving the problem. It also serves to prevent an increase in density for similar terms that originate from the seed problem yielding good problem samples for language models \cite{lee2021deduplicating}.
    \item \textbf{Added unrelated information} ---
        Some variations contain an extra phrase or quantity, or similar additions that are in excess of the information required to solve a problem and do not affect the original problem formulation in any meaningful way. These adversarial variations serve to obfuscate and familiarize the models with only the necessary information, enhancing deductive abilities \cite{kumar2021adversarial}.
    \item \textbf{Inverted question} ---
        Some variations will take a previously known quantity and turn it into an unknown quantity while revealing the previous unknown quantity of the problem. This, in many cases, alters the question drastically, changing the needed calculations and equations, while keeping a roughly similar question body to the seed problem. \citet{liu2021roda} used such problem samples in their work.
\end{itemize}
    \subsubsection{Seed Problems}
    Many of the seed problems used to generate variations from \textsc{Mawps} pose sufficient difficulty to even SOTA MWP solvers and often contain numeric information embedded within the statement itself. An example is the following problem, 
    \begin{quoting} 
    "\textit{Mary, Sam, Keith, and Alyssa each have 6 marbles. How many marbles do they have in all?}"
    \end{quoting}
    This problem yields the equation "$x=4\times6$", despite the quantity 4 not being mentioned anywhere in the statement. This quantity had to be inferred from the other parts of the statement itself, namely, the 4 entities referred to in the statement; Mary, Sam, Keith, and Alyssa. Another such problem is,
    \begin{quoting}
    "\textit{When the price of diesel rose by 10\%, a user reduced his diesel consumption by the same amount. How much would his diesel bill change in terms of percentage?}"
    \end{quoting}
    which yields the complex equation of "$x=(1.0-((1.0+(10.0\times0.01))\times(1.0-(10.0\times0.01))))\times100.0$". This problem, although seemingly simple on the surface in terms of quantities described, has several calculations dictated through the problem statement, some of which require additional real-world anecdotal knowledge, such as the conversion of percentages. Another problem with similar inferences of a more complex nature is, 
    \begin{quoting}
    "\textit{Lauren wants to mix 5 liters of 7\% milk with skim-milk (0\% fat) to produce a mixture of 2.9787\% milk. How much skim-milk should Lauren add?}"
    \end{quoting}
    yielding the equation "$x=(7.0\times0.01)\times5.0/(2.9787\times0.01)-5.0$", containing similar conversions of percentages, as well as additional knowledge of types of mixtures. Here, 7\% milk is mixed with pure milk, or 100\% milk. Yet the only indication that the milk is of 100\% purity is nowhere to be seen in a direct capacity in the problem, but rather in a roundabout way - by referring to the amount of fat (0\%) rather than the purity of the milk. Models have to infer a vast amount of real-world contextual knowledge to be able to solve such problems. Problems with second-degree unknown quantities are also present as seed problems. For example, the problem
    \begin{quoting}
    "\textit{The Hudson River flows at a rate of 3 miles per hour. A patrol boat travels 60 miles upriver and returns in a total time of 9 hours. What is the speed of the boat in still water?}"
    \end{quoting}
    that yields the equation "$(60.0/(x-3.0))+(60.0/(3.0+x))=9.0$", which is a quadratic equation. The problem itself deals with calculations of speed, which requires knowledge of how speed is calculated given certain quantities, as well as the effect of certain elements in the problem scenario on speed.

We resort to this data generation approach due to the lack of large-scale, diverse, single-equation English MWP datasets. Other commonly-used benchmark datasets, \textsc{Math23K} \cite{wang2017deep} and \textsc{Ape210K} \cite{liang2021mwp} consist of math problems written in Chinese Mandarin. We also aim to diversify the samples in \textsc{Mawps} to enable better training for MWP solvers \cite{schick2021generating,kumar2022practice}. \textsc{Svamp}, created by \citet{patel2021nlp} consists of challenging versions of problems and is considered a challenge set for testing the robustness of MWP solvers. We use the original version of \textsc{Mawps} and \textsc{Svamp} along with our dataset \textsc{ParaMAWPS} for conducting our experiments. A comparative summary of the statistics of the datasets used is shown in Table-\ref{tab:compare-datasets} and their operator count distributions are portrayed in Figure-\ref{fig:fig2}.
\begin{table}[h]
    \scriptsize
    \centering
    \begin{tabular}{c|c|c|c}
    \hline
    \textbf{Properties} & \textsc{Svamp} & \textsc{Mawps} & \textsc{ParaMAWPS} \\ \hline
    \# of problems & 1,000 & 2,373 & 16,278 \\
    \# of unique templates & 27 & 159 & 215 \\
    Avg. \# of operators & 1.236 & 1.606 & 1.68 \\
    Avg. \# of quantities per prob. & 2.81 & 2.57 & 2.54 \\
    Avg. \# of quantities per equ. & 2.23 & 2.59 & 2.67 \\
    \# of problems with constants & 0 & 185 & 3313 \\ \hline
    \end{tabular}
    \caption{Comparison of the datasets used.}
    \label{tab:compare-datasets}
    \end{table}
    \vspace{-2mm}
    \subsection{Model Implementation Details and Training}
    \subsubsection{Baseline Models}
    \label{baseline_models}
    We implement the DeBERTa model using Microsoft's \textit{deberta-base} that is publicly available in Hugging Face\footnote{\fontsize{8pt}{3pt} \selectfont{\url{https://huggingface.co/microsoft/deberta-base}}}. The other baseline MWP solver models are implementations already available in the open-source \texttt{MWPToolkit}\footnote{\fontsize{8pt}{3pt} \selectfont\url{https://github.com/LYH-YF/MWPToolkit}} developed by \citet{lan2022mwptoolkit}. We use an extensive set of baseline models, Transformer \cite{vaswani2017attention}, DNS \cite{wang2017deep}, MathEN \cite{wang2018translating}, GroupATT \cite{li2019modeling}, RNNEncDec \cite{sutskever2014sequence}, RNNVAE \cite{su2018variational}, BERT \cite{jdevlin2018bert}, RoBERTa \cite{liu2019roberta}, and compare them with the performance of the DeBERTa model. See Appendix-\ref{sec:appendix} for more training process details.


    \subsection{Result Analysis}
    \begin{table}[ht]
    \centering
    \footnotesize
\begin{tabular}{l|c|c|C{1cm}}
\hline
\textbf{Methods}            & \begin{tabular}{c}\scriptsize{\textsc{Mawps}\textsuperscript{\textdagger}}\\ (\%)\end{tabular} & \begin{tabular}{c}\scriptsize{\textsc{Svamp}}\\ (\%)\end{tabular} & \scriptsize{\textsc{ParaMawps}}\textsuperscript{\textdagger} (\%)  \\ \hline
DNS                & 59.5       & 22.1 & 71.2\\
Math-EN            & 69.2       & 21.8 & 71.6\\
GROUP-ATT          & 76.1       & 19.2 & 70.8\\
RNNEncDec              & 79.4       & 25.4 & 73.6\\
RNNVAE              & 79.8       & 25.9 & 72.8\\
Transformer & 85.6       & 20.7 & 64.6\\
BERT     & 86.9   & 24.8 & 72.1      \\ 
RoBERTa     & 88.4   & 30.3 & 72.5      \\ \hline
DeBERTa     & 90.7   & \textbf{63.5} & 74.1\\
DeBERTa\textsubscript{\tiny{PM + VM}}     & \textbf{91.0}   & - & -\\
DeBERTa\textsubscript{\tiny{VM}}     & -   & - & \textbf{79.1}\\
\hline
\end{tabular}
\caption{Value accuracy of the DeBERTa model and various baseline models. \textdagger \vspace{1pt} denotes 5-fold cross validation. PM stands for Paraphrasing Model and VM stands for Voting Mechanism.}
\label{tab:comparison}
\end{table}
Table-\ref{tab:comparison} shows the performance comparison of the DeBERTa model and the baseline models mentioned in Section-\ref{baseline_models}. The DeBERTa model coupled with the Paraphrasing model and the Voting Mechanism outperforms all the baseline models in the \textsc{Mawps} \cite{koncel2016mawps} dataset with an accuracy of $91.0\%$. The Paraphrasing Model and the Voting Mechanism contributed to a $0.3\%$ increase in accuracy. The vanilla DeBERTa model also outperforms the baseline models in our \textsc{ParaMAWPS} dataset by boasting an accuracy of $74.1\%$. With the voting mechanism at the tail-end of the pipeline, we are able to yield an improvement of the accuracy by $5.04\%$ making the accuracy $79.1\%$. We test the robustness of the vanilla DeBERTa model on the \textsc{Svamp} \cite{patel2021nlp} challenge dataset and get an accuracy of $63.5\%$ which is quite higher than that of the other baseline models. The model still lags a mere $1 \pm 0.20\%$ behind the current SOTA model on \textsc{Mawps}, which is the \textsc{RoBERTa-DeductReasoner} model by \citet{jie2022learning} ($92.0 \pm 0.20\%$) but supersedes its accuracy of $47.3\pm0.20\%$ on the \textsc{Svamp} dataset. 

The superiority of the model's accuracy in \textsc{ParaMAWPS} over \textsc{Svamp}, despite the demonstrably greater difficulty of the MWP samples in \textsc{ParaMAWPS}, indicates that training a language model on a more diverse set of linguistically varied problem statements leads to a better quality mathematical reasoning ability after the training phase.

\subsection{Ablation Study}
To gain insights into the individual contributions of the Paraphrasing Model and Voting Mechanism in conjunction with the DeBERTa model, we perform ablation studies.
\begin{table}[h]
\footnotesize
\centering
\begin{tabular}{l|l}
\textbf{\# of variants} & \textsc{Mawps}\textsuperscript{\textdagger} (\%) \\ \hline
w/ $k=0$ & 90.7 \\ \hline
w/ $k=5$ & 90.4 \\ \hline
w/ $k=10$ & 90.8 \\ \hline
w/ $k=15$ & 91.0 \\ \hline
\end{tabular}
\caption{Value accuracy with different numbers of linguistic variants of the problem samples. \textdagger \vspace{1pt} denotes 5-fold cross validation.}
\label{tab:ablation1}
\end{table}
\begin{table}[h]
\footnotesize
\centering
\begin{tabular}{l|l}
\textbf{Voting Mechanism} & \textsc{ParaMawps}\textsuperscript{\textdagger} (\%) \\ \hline
w/o VM & 72.9, 74.1, 76.5, 72.1, 74.6 \\ \hline
w/ VM & 78.5, 77.8, 82.4, 77.2, 79.5 \\ \hline
\end{tabular}
\caption{Effect of Majority Voting on Value accuracy across all 5 folds. \textdagger \vspace{1pt} denotes 5-fold cross validation.}
\label{tab:ablation2}
\end{table}
\\
Table-\ref{tab:ablation1} shows the effect of increasing the number of generated problem variants to infer the solution expressions of the problem samples in the \textsc{Mawps} dataset's test set. Although there is a slight decrease in the accuracy for $k=5$, we see a minuscule increase in accuracy for $k=10$ and $k=15$. In Table-\ref{tab:ablation2} we see the impact of the Voting Mechanism which contributed to a $5.4\%$ increase on average in the accuracy of the DeBERTa model on the \textsc{ParaMAWPS} dataset.

\subsection{MWP Task Performance Analysis of Large Language Models}
    \label{sec:mwpllm}
    To test out the assertion made in other studies \cite{huang2022towards, ho2022large} about the incompetence of LLMs in complex reasoning tasks compared to fine-tuned smaller models, we use the GPT-J model and some of the presently used GPT-3 models by OpenAI to perform the task of MWP solving. We use the original version of \textsc{Mawps} \cite{koncel2016mawps} along with our dataset \textsc{ParaMAWPS} for testing the mathematical reasoning of these models. 
    \begin{table}[h]
    \centering
\begin{tabular}{l|c|C{1cm}}
\hline
\textbf{Models}            & \begin{tabular}{c}\small{\textsc{Mawps}\textsuperscript{\textdagger}}\\ (\%)\end{tabular} & \footnotesize{\textsc{ParaMawps}}\textsuperscript{\textdagger} (\%)  \\ \hline
GPT-J (6B)               & 9.9       & 5.9\\
\textit{text-babbage-001} (6.7B)          & 2.76       & 3.21\\
\textit{text-curie-001} (13B)         & 4.09       & 4.20\\
\textit{gpt-3.5-turbo} (175B)             & 80.3      & 73.0\\ 
\hline
\end{tabular}
\caption{Value accuracy of the LLMs in a zero-shot setup testing. \textdagger \vspace{1pt} denotes evaluation on the whole dataset.}
\label{tab:llms}
\end{table}
\\
    One of the most capable models in the GPT-3.5 series of models is \textit{text-davinci-003}, with 175 billion parameters and the ability to follow instructions consistently and produce lengthy outputs. However, the most capable and up-to-date model according to OpenAI is \textit{gpt-3.5-turbo}, with 175 billion parameters, which is primarily optimized for chat completions but can be tweaked to follow more specific instructions similar to \textit{text-davinci-003}. While all models used are instructed to output in a specific format --- \texttt{`Answer: [ANS]'} with just the numerical value in the place of \texttt{`[ANS]'}, the ability to do so consistently deteriorated with the models with relatively fewer parameters. Out of the base GPT-3 models, the 13 billion parameters \textit{text-curie-001} can yield outputs in the given format relatively consistently and \textit{text-babbage-001}, with 6.7 billion parameters can occasionally produce the output in the correct format, but tries to generate full sentences more often than not, whereas the 350 million parameters \textit{text-ada-001} can barely generate a single output in the correct format, choosing to generate full sentences almost all of the time. Models tend to try to \textit{`work through'} the problem in text form rather than just generating the output, although with \textit{gpt-3.5-turbo} this can be mostly mitigated by using very specific instructions for the prompt. The results in Table-\ref{tab:llms} and Table-\ref{tab:comparison} support the current weakness of LLMs in mathematical reasoning tasks and the suitability of fine-tuning smaller models. It indicates the improvement in performance for a well-reasoning, but comparatively small model when it has the option to democratically choose from a substantial number of solution guesses. 
\section{Conclusion and Future Work}
In this paper, we propose the idea of an MWP solving framework that utilizes the paraphrased linguistic variations of problem texts to train a DeBERTa model that generates candidate solution expressions and finalizes the predicted math expression by employing majority voting on a set of simplified candidate expressions. Our findings demonstrate that incorporating linguistic variants of problem statements during training and utilizing a voting mechanism for candidate predictions enhance the model's mathematical reasoning and overall robustness.

We also introduce a large-scale, diverse, and challenging single-equation MWP dataset, \textsc{ParaMawps}, consisting of paraphrased, inverse, and adversarial variants of selectively sampled datapoints from \textsc{Mawps}, as a formidable evaluation test-bed and a proper benchmark for training MWP solver models.

We wish to experiment further with harder problem text variations (\textit{e.g.} grammatical errors) and conduct a thorough error analysis of the models for identifying their lapses in mathematical reasoning and discovering more scopes of improvement. We also aim to expand our research to encompass the intricate realms of multi-equation, multi-step deduction, and domain-knowledge problems. We hope our approach and findings will pave the way to more scholarly works on the vistas of AGI and in tandem be deemed a noteworthy and meaningful contribution to this domain of research.
\section{Limitations}
There are still some avenues of improvement in our work. The temporal overhead due to the problem variant generation by the paraphraser model may make our proposed architecture unsuitable for real-world applications even though it takes merely 10 to 12 seconds to generate $k=5$ variants for a single sample. 
Another limitation of our work is the absence of a proper tie-breaking strategy in our Majority Voting module. Furthermore, we need to introduce a system of weighted votes (\textit{e.g.} semantic similarity scores as weights) so that the votes of wrongly predicted equations don't trump that of correctly generated predictions. We also plan to incorporate and experiment with the Tree-based decoder \cite{xie2019goal} in our proposed pipeline.

\section*{Acknowledgements}
We convey our heartfelt gratitude to the anonymous reviewers and the mentors of the pre-submission mentorship program for their constructive criticisms and insightful feedback which were conducive to the improvement of the research work outlined in this paper. We also appreciate the Systems and Software Lab (SSL) of the Islamic University of Technology (IUT) for the generous provision of computing resources during the course of this project. Syed Rifat Raiyan, in particular, wants to thank his parents, Syed Sirajul Islam and Kazi Shahana Begum, for everything.

\newpage
\bibliography{anthology,custom}
\bibliographystyle{acl_natbib}

\appendix

\section{Appendix}
\label{sec:appendix}
    \subsection{Dataset Split}
    We use an $80$:$10$:$10$ train-validation-test split for our \textsc{ParaMAWPS} dataset. For \textsc{Mawps}, we use 5-fold cross-validation using the splits provided by its authors \citet{koncel2016mawps}. The \textsc{Svamp} dataset is a challenge set and all 1,000 of its samples constitute the test set while the model itself is trained on a combination of the \textsc{Mawps} and \textsc{ASDiv-A} \cite{miao2021diverse} dataset.

    \subsection{Performance Evaluation and Metric}
    We use Negative log-likelihood loss (NLLLoss) for training all the models. For the baseline models, \texttt{MWPToolkit} uses two metrics of accuracy, \textit{Equation Accuracy} and \textit{Value Accuracy}. Equation accuracy measures the correctness of the generated equation. Value accuracy measures the correctness of the value yielded from evaluating the generated equation. This metric takes into consideration the fact that models may generate equations that have a different template than the respective ground truth equations but nevertheless yield the correct answers to the problem statements. 
    
    \subsection{Hyperparameters}
    In the DeBERTa model, we use embedding dimension $d = 768$, $FFN_{size} = 1024$, number of decoder layers $N = 4$, number of attention heads $h = 16$, dropout ratio $P_{drop} = 0.5$, learning rate $lr = 10^{-5}$, batch size $b = 8$, and $Epochs = 200$. The hyperparameters for the other baseline models are as set on the respective \texttt{MWPToolkit} implementations.

    \subsection{Optimizer}
    We use Adam \cite{kingma2014adam} with a StepLR learning rate scheduler as our optimizer. The learning rate $lr$ is set according to \citet{vaswani2017attention}, $lr = d^{-0.5} \cdot \operatorname{min}(n^{-0.5},n \cdot w^{-1.5})$
    where, $d$ is the embedding dimension, $n$ is the step number and $w$ is the number of warm-up steps. Here, warm-up steps $w$ simply insinuate that the learning rate rises linearly for the initial $w$ training steps. We set $\beta_1 = 0.9$, $\beta_2 = 0.999$, $\epsilon = 10^{-8}$ and $w = 1500$ for the models' Adam optimizer. For the StepLR scheduler, we set $\gamma=0.5$ and $step\_size = 5$.
    
    \subsection{Hardware and Schedule}
    We have used the NVIDIA RTX 3090 GPU equipped with 25GB of VRAM and Intel Core i9 Processor for conducting our experiments. The DeBERTa model took around 18 hours to fully train on the \textsc{ParaMAWPS} dataset with 5-fold cross-validation and 200 epochs per fold, which was the highest expense of time among the lot. The other baseline models took approximately 7 to 9 hours on the \textsc{ParaMAWPS} dataset and around 5 hours on \textsc{Mawps} and \textsc{Svamp}. The greater the number of parameters that a model possesses the more time it takes to fully complete the 5-fold training process. As DeBERTa has an astounding 134 million parameters \cite{he2020deberta}, it takes the longest time to train.

    \definecolor{NavyBlue}{HTML}{0066CC}
    \definecolor{ForestGreen}{HTML}{32CD32}
    \definecolor{Lavender}{HTML}{745BB1}
    \definecolor{Crimson}{HTML}{DC143C}
\begin{table*}
        \centering
        \footnotesize
    \resizebox{\textwidth}{!}{%
    \begin{tabular}{C{6em}m{20em}m{20em}}
    \toprule
    \textbf{Variation Type} & \textbf{Original} & \textbf{Variation} \\ \hline
    Changed phrase order & There were originally 20817 houses in Lincoln County. During a housing boom, developers built 97741. How many houses are there now in Lincoln County? & \textcolor{NavyBlue}{How many houses are there in Lincoln County now, after developers built an additional 97741 during a housing boom, when there were originally 20817 houses?} \\ \hline
    Changed object and entity names & While playing a \textcolor{cyan}{trivia game}, \textcolor{cyan}{Mike answered} 3 \textcolor{cyan}{questions} correct in the first half and 5 \textcolor{cyan}{questions correct} in the second half. If each \textcolor{cyan}{question} was worth 3 points, what was \textcolor{cyan}{his} final score? & While playing a \textcolor{ForestGreen}{game of Hangman}, \textcolor{ForestGreen}{Emily guessed} 3 \textcolor{ForestGreen}{letters correctly} in the first half and 5 \textcolor{ForestGreen}{letters correctly} in the second half. If each \textcolor{ForestGreen}{letter} was worth 3 points, what was \textcolor{ForestGreen}{her} final score? \\ \hline
    Added unrelated information & A carpenter bought a piece of wood that was 8.9 centimeters long. Then he sawed 2.3 centimeters off the end. How long is the piece of wood now? & A carpenter bought a piece of wood that was 8.9 centimeters long. Then he sawed 2.3 centimeters off the end \textcolor{red}{and sanded the wood for 20 minutes}. How long is the piece of wood now? \\ \hline
    Inverted question & Mary bought 3 pizzas for \$8 each. What was the total amount she paid for the 3 pizzas? & If Mary paid \textcolor{violet}{\$24} for 3 pizzas, \textcolor{violet}{how much did she pay for each pizza}? \\ \bottomrule
    \end{tabular}%
    }
    \caption{Types of Variations with examples. The problems in the \textbf{Original} column are samples taken from the \textsc{Mawps} dataset, whereas, the ones in the \textbf{Variation} column are from the \textsc{ParaMAWPS} dataset.}
    \label{tab:variations}
    \end{table*}
\begin{figure*}
        \centering
        \includegraphics[scale=0.14]{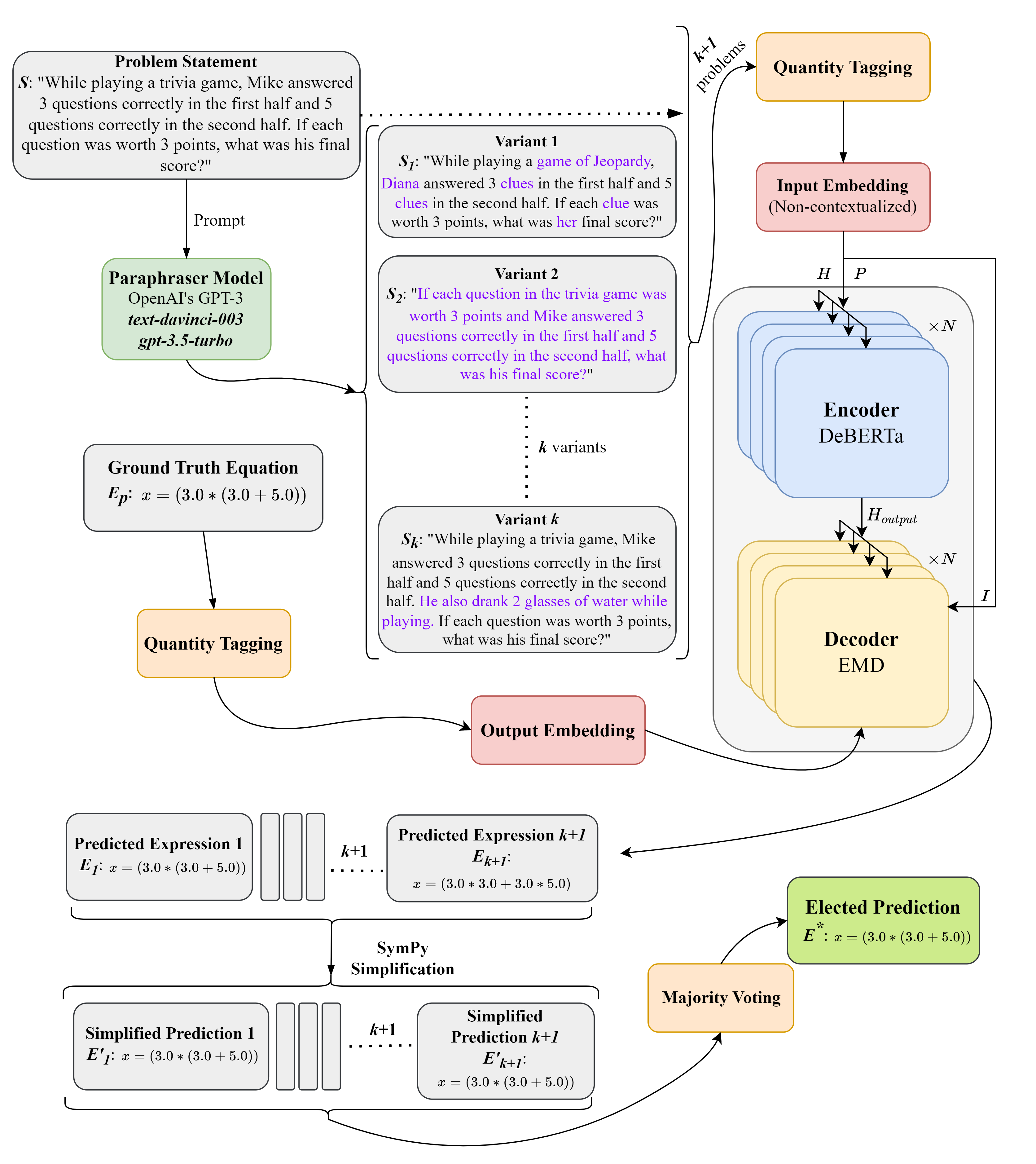}
        \caption{Overview of our proposed architecture.}
        \label{fig:proposedmodel}
\end{figure*}
\begin{figure}[h]
    \captionsetup[subfigure]{labelformat=empty}
    \centering
    \subfloat[]{\includegraphics[scale=0.33]{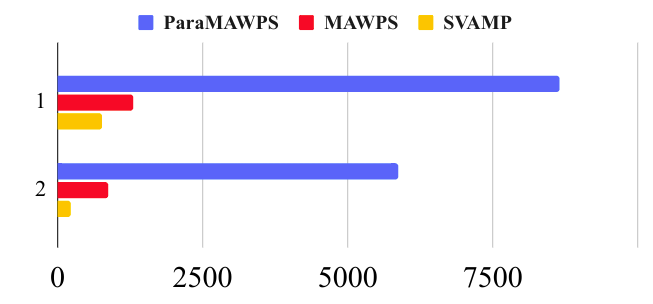}}\hspace{0.0cm}\subfloat[]{\includegraphics[scale=0.33]{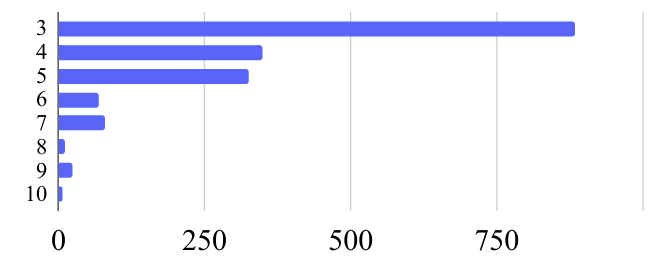}}\hspace{0.0cm}\subfloat[]
    {\includegraphics[scale=0.33]{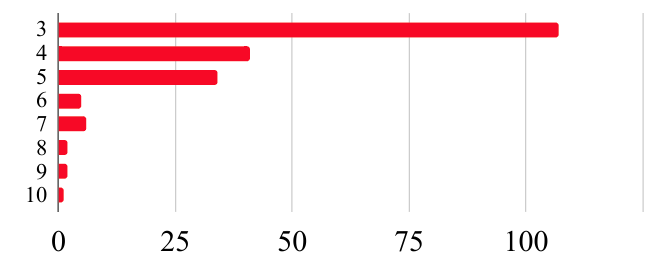}}
    \caption{Operator count distributions of \textsc{ParaMAWPS}, \textsc{Mawps}, and \textsc{Svamp}. We keep the distribution of \textsc{ParaMAWPS} somewhat similar to that of \textsc{Mawps} to maintain a proper balance between easy and difficult problems.}
    \label{fig:fig2}
    \end{figure}
\end{document}